\documentclass[preprint,12pt]{elsarticle}

\usepackage{algorithm}
\usepackage{algorithmic}
\usepackage{graphicx}
\usepackage{multirow}
\usepackage{makecell}
\usepackage{booktabs}
\usepackage{float}
\usepackage{graphicx}
\usepackage[switch]{lineno}
\usepackage{amssymb}
\usepackage{array}
\usepackage{hyperref} 
\usepackage{color}
\usepackage{amsthm}
\usepackage{lineno}
\usepackage{amsmath}

\biboptions{numbers,sort&compress}

\journal{Pattern Recognition}

\begin{document}

\begin{frontmatter}

% \linenumbers


\title{Towards Local Visual Modeling for Image Captioning}

\author[xmu]{Yiwei Ma}
\author[xmu]{Jiayi Ji}
\author[xmu,xmuai]{Xiaoshuai Sun}
\author[xmu]{Yiyi Zhou}
\author[xmu,xmuai,pengcheng]{Rongrong Ji}

\address[xmu]{Media Analytics and Computing Lab, Department of Artificial Intelligence, School of Informatics, Xiamen University, 361005, China}
\address[xmuai]{Institute of Artificial Intelligence, Xiamen University}
\address[pengcheng]{Peng Cheng Laboratory, Shenzhen, China}

% \author{}

% \affiliation{organization={},%Department and Organization

\begin{abstract}
In this paper, we study {\color{black} the} local visual modeling with grid features for image captioning, which is critical for generating accurate and detailed captions.  To achieve this target, we propose a  \emph{Locality-Sensitive Transformer Network} (LSTNet) with two novel designs, namely \emph{Locality-Sensitive Attention} (LSA) and \emph{Locality-Sensitive Fusion} (LSF). LSA is deployed for the intra-layer interaction in Transformer via modeling the relationship between each grid and its neighbors. It reduces the difficulty of local object recognition during captioning. LSF is used for inter-layer information fusion, which aggregates the information of different encoder layers for cross-layer semantical complementarity. With these two novel designs, the proposed LSTNet can model the local visual information of grid features to improve the captioning quality. To validate LSTNet, we conduct extensive experiments on the competitive MS-COCO benchmark. The experimental results show that LSTNet is not only capable of local visual modeling, but also outperforms a bunch of state-of-the-art captioning models on offline and online testings, \emph{i.e.,} 134.8 CIDEr and 136.3 CIDEr, respectively. Besides, the generalization of LSTNet is also verified on the Flickr8k and Flickr30k datasets. The source code is available on GitHub: {\url{https://github.com/xmu-xiaoma666/LSTNet}}.
\end{abstract}


% \begin{highlights}
% \item Local visual modeling with grid features for image captioning.
% \item Locality-Sensitive Attention (LSA) is deployed for the intra-layer interaction in Transformer via modeling the relationship between each grid and its neighbors.
% \item Locality-Sensitive Fusion (LSF) is used for inter-layer information fusion, which aggregates the information of different encoding layers for the cross-layer semantical complementarity.
% \item Locality-Sensitive Transformer Network (LSTNet) outperforms a bunch of SOTA captioning models on the competitive MS-COCO benchmark.
% \item The generalization of LSTNet is also verified on the Flickr8k and Flickr30k datasets.
% \end{highlights}


\end{frontmatter}



% \linenumbers

\section{Introduction \label{sec:intro}}

Image captioning is the task of generating a fluent sentence to describe the given image. Recent years have witnessed the rapid development of this field, which is supported by a  bunch of innovative methods~\cite{cornia2019show,cornia2020meshed} and datasets~\cite{lin2014microsoft,plummer2015flickr30k}.
% LI202168,LIU2021187,WAN2022108358,JI2021107928,YANG2022108545,LIM2022108285,ZHOU202134,TAN2022108366,

Inspired by the great success of \emph{Bottom-up Attention} \cite{anderson2018bottom}, most existing methods in image captioning adopt the region features extracted by the object detector as the visual representations, \emph{e.g.,} Faster R-CNN \cite{ren2016faster}. Since the detector is pre-trained on the large-scale Visual Genome dataset \cite{krishna2017visual}, it can generate discriminative representations for salient regions in the image and provide complete object information for captioning. To this end, significant progress in image captioning has been made based on the region features \cite{pan2020x,cornia2020meshed,huang2019attention}.

\begin{figure}[ht]
\centering 
  \includegraphics[width=0.7\columnwidth]{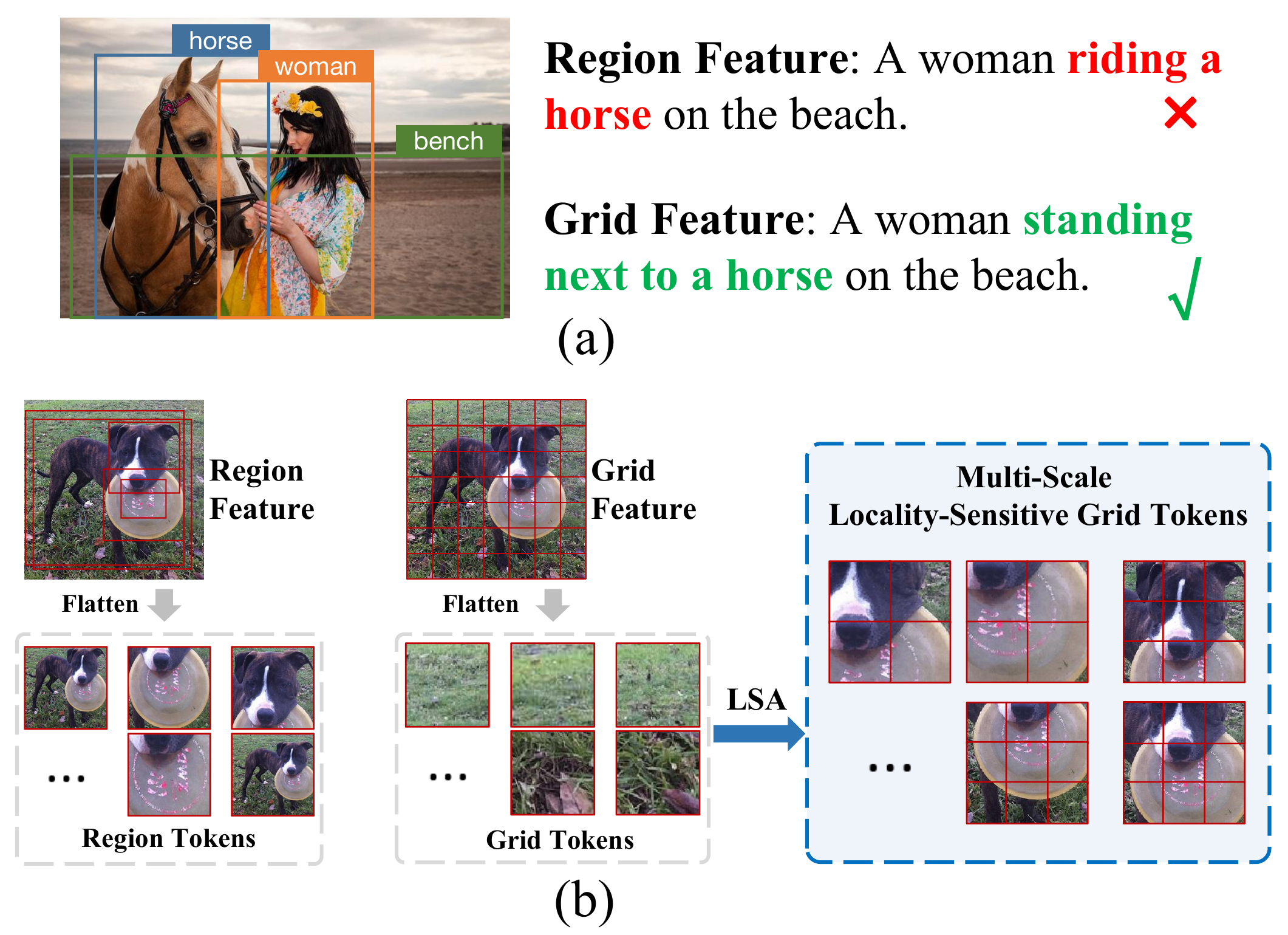}
  \vspace{-0.00cm}
{\color{black}{
  \caption{(a)  Captions generated by Transformer with the region and grid features, respectively. 
  (b)  Region features often contain complete object information, while the grid ones are more fragmented. Our LSA is conducive to reconstructing complete object information by modeling the relationship of adjacent grids. 
  }
  \label{fig:fig1}
}}
  \vspace{-0.00cm}
\end{figure}

However, the region features still {\color{black}{exist}} obvious defects. To be specific, {\color{black}{they}} are extracted from the salient regions of the image, thus often ignoring the contextual information {\color{black}{in the background}}. In this case, {\color{black}{it is inferior for the model to capture the relationship between objects}}. For example, as shown in Fig.~\ref{fig:fig1}(a), the model trained on region features fails to understand the contextual information out of the bounding boxes, thereby incorrectly describing the relationship between \emph{``woman"} and \emph{``horse"}. Besides, the pre-trained object detector {\color{black}{may produce}} noisy, overlapped,  or erroneous detections, which ultimately limits the performance {\color{black}{upper bound}} of image captioning models. 

To compensate for the aforementioned limitations, {\color{black}{some}} endeavors start to revisit the use of grid features. \cite{jiang2020defense} {\color{black}{explores}} grid features of {\color{black}{the}} object detector to further push the performance of the visual question answering (VQA) task. RSTNet \cite{zhang2021rstnet} and DLCT \cite{luo2021dual} first adopt grid features in Transformer-like networks, which achieve impressive performance in image captioning. However, Transformer-like architectures are not conducive to the perception of complete objects. Specifically, as shown in Fig.~\ref{fig:fig1}(b), {\color{black}{a complete object may be divided into multiple adjacent grids in 2D space, while the flatten operation in Transformer inevitably destroys the local relationship of grid features}}. Meanwhile, recent advances \cite{wu2021cvt} also show that the vanilla Transformer is less efficient in local visual modeling. 

{\color{black}{
Based on the above analysis, we observe that both region and grid features have their own advantages and disadvantages. Region features contain explicit object information but lack background and relationship information. In contrast, grid features contain all information at the same time, but an object may be divided into multiple grids. As a result, the majority of the semantic information is damaged, which makes reasoning more challenging. A straightforward solution to enjoy the benefits of both features is adopting both region and grid features like DLCT~\cite{luo2021dual} and GRIT~\cite{nguyen2022grit}. However, it will lead to significantly higher computation and longer training time, because the model needs to process both features at the same time. A more efficient way is to model the local information on the grid features to compensate for the lack of object information.
}}
% To this end, we argue that only applying self-attention to model the dependency of grid features is not sufficient for local visual recognition, {\color{black}{leading}} to sub-optimal captioning performance. 

{
\color{black}{Therefore, we propose a novel \emph{\textbf{Locality-Sensitive Transformer Network (LSTNet)}} in this paper. Specifically, LSTNet strengthens local modeling to perceive object-level information from the aspects of \emph{intra-layer interaction} and \emph{inter-layer fusion}, respectively.}} For intra-layer interaction, we propose a novel multi-branch module called \emph{\textbf{Locality-Sensitive Attention (LSA)}} to perceive fine-grained local information from different receptive fields and enhance the interactions {\color{black}{between}} each grid and its neighbors. Notably, LSA can be re-parameterized into a single-branch structure during {\color{black}{inference}}, thereby reducing the {\color{black}{additional}} overhead of multi-scale {\color{black}{perception}}.  For inter-layer fusion, we {\color{black}{design}} a \emph{\textbf{Locality-Sensitive Fusion (LSF)}} module, which can align and fuse grid features from different layers for cross-layer semantical complementary. 
With these novel designs, LSTNet {\color{black}{improves the ability of}} of local visual modeling, but also greatly improves the quality of {\color{black}{the generated}} captions.
On the competitive MS-COCO benchmark, LSTNet presents outstanding {\color{black}{performance}} on both offline and online testing, \emph{i.e.,} 134.8 CIDEr and 136.3 CIDEr. In addition to the outstanding performance on the MS-COCO dataset, the generalization of LSTNet is also verified on the Flickr8k and Flickr30k datasets.

To sum up, our contributions are three-fold:
\begin{itemize}
    \item {\color{black}{To perceive object and context information with only grid features, we propose a novel LSTNet for image captioning.}} LSTNet {\color{black}{not only improves the local perception ability of the model}} but also outperforms a bunch of recently proposed methods on the highly competitive MS-COCO benchmark. %\footnote{ For fair comparisons, we do not compare methods using large-scale multi-modal pre-training.}

    \item We propose a \emph{Locality-Sensitive Attention (LSA)} for the intra-layer visual modeling in Transformer, which is  a re-parameterized module for enhancing the interaction between each grid feature and its local neighbors.  
    
    \item We propose a \emph{Locality-Sensitive Fusion (LSF)} to aggregate inter-layer object semantic information for image captioning, which is conducive to  inter-layer semantic understanding.
    
\end{itemize}

\begin{figure}[ht]
\centering 
  \includegraphics[width=0.8\columnwidth]{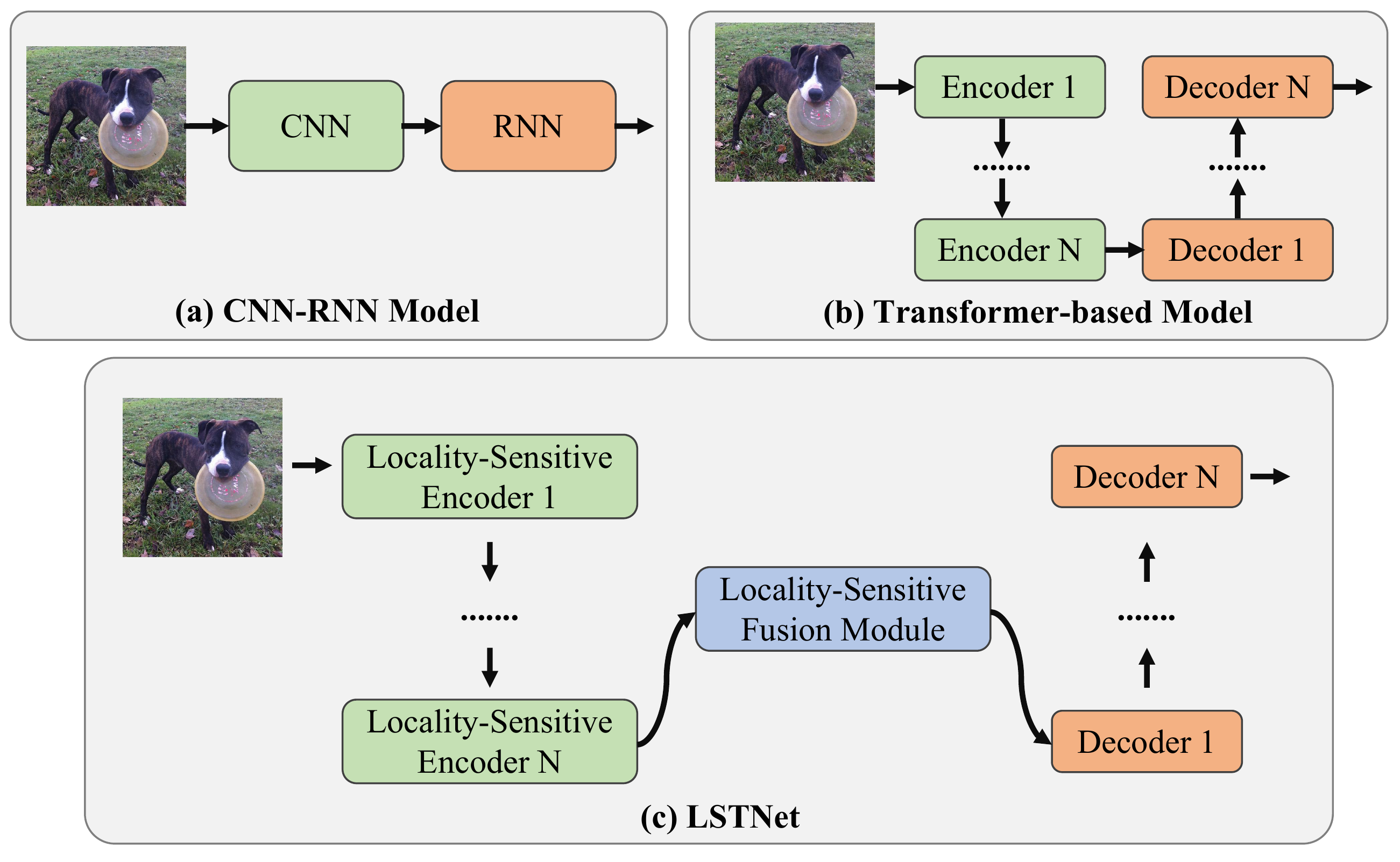}
  \vspace{-0.00cm}
{\color{black}{
  \caption{Illustration of CNN-RNN model (a), Transformer-based model (b), and LSTNet (c) for image captioning.}
  \label{fig:model}
}}
  \vspace{-0.00cm}
\end{figure}

\section{Related Work}

\subsection{Image Captioning}

Image captioning is a challenging task, and enormous effort has been made to solve this problem. With years of development,  a great improvement can be observed with a flurry of methods \cite{vinyals2015show,lu2017knowing,anderson2018bottom,pan2020x,zhang2021rstnet,ji2022knowing,ma2022knowing}. {\color{black}{The existing image captioning methods can be roughly divided into two categories: 1) the CNN-RNN model, 2) the Transformer-based model. As shown in Fig. \ref{fig:model}(a), the CNN-RNN model uses CNN to encode images into vectorial representations and then adopts an RNN-based decoder to fuse these vectorial representations to provide content-related descriptions for input images. Specifically, \cite{vinyals2015show} uses Convolutional Neural Network (CNN) to encode images and adopts Long Short-Term Memory (LSTM) as a decoder to generate captions. \cite{lu2017knowing} exploits the adaptive attention mechanism to decide whether to attend to visual or non-visual information at each time step. \cite{anderson2018bottom} uses {\color{black}{the}} pre-trained Fast R-CNN \cite{ren2016faster} to extract salient objects as regional visual features, which is conducive to generating accurate captions. With the development of Transformer \cite{vaswani2017attention}, a lot of researchers are investigating the application of Transformer-based models on the image captioning task, which is illustrated in Fig. \ref{fig:model}(b). \cite{pan2020x} introduces Bi-linear Pooling into the Transformer model to capture $2^{nd}$ order interactions. \cite{zhang2021rstnet} proposes to adaptively measure the contribution of visual and language cues on the top of the transformer decoder before word prediction. \cite{luo2021dual} presented a Dual-level Collaborative Transformer (DLCT) to accomplish the complementarity of the region and grid features.}} {\color{black}{To improve the semantic understanding ability of Transformer, \cite{ma2022knowing} proposes a Transformer-based captioning model with both spatial and channel-wise attention. \cite{WANG2022117174}} proposes a Geometry Attention Transformer (GAT) model to further leverage geometric information in image captioning. To consider the visual persistence of object features, \cite{WANG202248} introduces a VPNet via inserting visual persistence modules in both the encoder and decoder. \cite{zhang2021exploring} proposes a novel CtxAdpAtt model, which adopts the linguistic context to explore related visual relationships between different objects effectively. To alleviate the disadvantages of using GCN-based encoders to represent the relation information among scene graphs, ReFormer\cite{yang2022reformer} explores a novel architecture to explicitly express the relationship between objects in the image.}

Our LSTNet is in line with the Transformer-base approach. However, when processing grid features, the Transformer ignores visual locality, which {\color{black}{is important for}} identifying objects in the image. As shown in Fig. \ref{fig:model}(c), we propose the Locality-Sensitive Attention (LSA) module and Locality-Sensitive Attention (LSA) module to enhance local visual modeling.

\begin{figure}[ht]
\centering 
  \includegraphics[width=0.5\columnwidth]{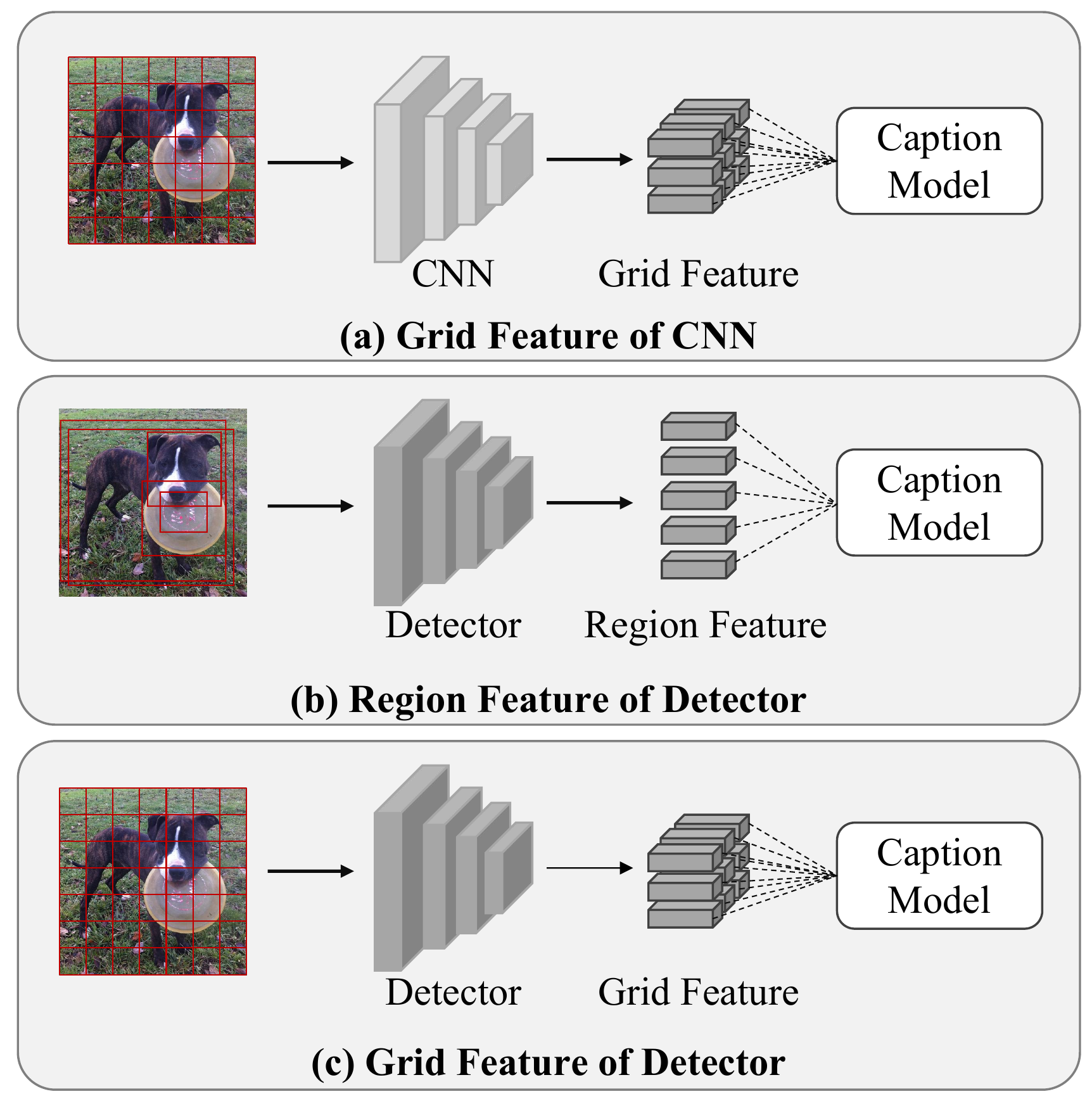}
  \vspace{-0.00cm}
{\color{black}{
  \caption{Three main stages of visual features used in {\color{black}{the}} image captioning task: \textbf{(a)} Grid features extracted by CNN, \emph{e.g.,} ResNet \cite{he2016deep}; \textbf{(b)} Region features extracted by the pre-trained detector, \emph{e.g.,} Faster R-CNN \cite{ren2016faster}, which requires time-consuming post-processing operations, \emph{e.g.,} NMS \cite{nms}; \textbf{(c)} Grid features extracted from the feature map of CNN in the object detector.}
  \label{fig:feature}
}}
  \vspace{-0.00cm}
\end{figure}

\subsection{Region features \& Grid features}
The visual features used in {\color{black}{image captioning}} go through three main stages: Grid $\rightarrow$ Region $\rightarrow$ Grid. In the first grid stage, some pioneering works \cite{xu2015show,vinyals2016show,lu2017knowing} adopt grid visual features extracted {\color{black}{from}} CNN \cite{he2016deep} to represent images, which is illustrated in Fig. \ref{fig:feature}(a).
{\color{black}{For example, \cite{xu2015show} first propose the image captioning task, and adopt CNN to encode visual features and RNN to decode the caption. To capture the importance of different grids in {\color{black}{an image}}, \cite{vinyals2016show} applies the attention mechanism to the visual features before decoding the caption.\cite{lu2017knowing} propose adaptive attention on the visual feature, which is extracted from the last convolutional layer of ResNet101~\cite{he2016deep}. }}
As shown in Fig. \ref{fig:feature}(b), to obtain foreground information, \cite{anderson2018bottom} {\color{black}{adopts}} an object detector \cite{ren2016faster} pre-trained on VG~\cite{krishna2017visual} to extract region features, which 
are widely used in a lot of multi-modal tasks \cite{cornia2020meshed,ma2022x}. As shown in Fig. \ref{fig:feature}(c), to compensate for the defects  (\emph{e.g.,} {\color{black}{time-consuming}}) of region features, \cite{jiang2020defense} {\color{black}{revisits}} the grid feature in the object detector, and {\color{black}{finds}} that it could achieve competitive performance in VQA, the effectiveness of which has also been validated in image captioning \cite{luo2021dual,zhang2021rstnet}.

Compared with previous methods \cite{huang2019attention,cornia2020meshed,pan2020x} based on region features, our proposed LSTNet based on grid features can capture the contextual information out of bounding boxes, thus generating more accurate captions. On the other hand, compared with existing methods \cite{zhang2021rstnet} based on grid features, our proposed LSTNet considers the locality of the grid features and models the relationship of neighboring grids, which is conducive to {\color{black}{recognizing}} the objects in the image. DLCT \cite{luo2021dual} adopts bounding boxes to assist grid features to locate objects. However, due to the adoption of both grid and region features, the model needs to bear more training and prediction overhead, \emph{e.g.,} LSTNet runs over three times faster than DLCT in the cross-entropy training stage, whose performance is severely limited by the accuracy of bounding boxes. Our proposed LSTNet captures intra- and inter-layer local relationships, leading to more detailed and finer-grained grid features for image captioning.

{\color{black}{
\subsection{Multi-head Self-Attention in Transformer}
Transformer {\color{black}{is}} originally proposed to solve natural language processing (NLP) tasks. Due to its powerful modeling ability, the transformer has also been widely used in computer vision (CV) and multi-modal tasks in recent years. The key component of the transformer is the multi-head self-attention (MSA) module, which can effectively model the relationship and context of the input at different positions. Specifically, a $h$-head self-attention {\color{black}{is}} formulated as: 

\begin{equation}
    \operatorname{MultiHead}(\mathbf{Q}, \mathbf{K}, \mathbf{V})=\text { Concat } \left({\text {head }_{1}}, \text { head }_{2}, \ldots, \text { head }_{\mathrm{h}}\right) \mathbf{W}^{\mathrm{O}},
\label{eq:msa1}
\end{equation}
where $\mathbf{Q}, \mathbf{K}, \mathbf{V} \in \mathbb{R}^{N \times d}$ represent the input query, key, and value, respectively. $N$ is the length of input, $d$ is the hidden dimension in each head. $\mathbf{W}^{O} \in \mathbb{R}^{h d \times d}$ is a learnable matrix for the output of all heads. For each head, the attention {\color{black}{is}} formulated as follows:

\begin{equation}
    \text { head }_{\mathrm{i}}=\operatorname{Attention}\left(\mathbf{Q}_{\mathrm{i}}^{\mathrm{Q}}, \mathbf{K} \mathbf{W}_{\mathrm{i}}^{\mathrm{K}}, \mathbf{V} \mathbf{W}_{\mathrm{i}}^{\mathrm{V}}\right)\\
    =\operatorname{softmax}\left(\frac{\mathbf{Q} \mathbf{W}_{\mathrm{i}}^{\mathrm{Q}}\left(\mathbf{K} \mathbf{W}_{\mathrm{i}}^{\mathrm{K}}\right)^{\mathrm{T}}}{\sqrt{\mathrm{d}}}\right) \mathbf{V} \mathbf{W}_{\mathrm{i}}^{\mathrm{V}},
\label{eq:msa2}
\end{equation}
where $\mathbf{W}_{\mathrm{i}}^{\mathrm{Q}}, \mathbf{W}_{\mathrm{i}}^{\mathrm{K}}, \mathbf{W}_{\mathrm{i}}^{\mathrm{V}} \in \mathbb{R}^{d \times d}$ are the learnable matrices for input query, key, and value, respectively.

}}

\section{Approach}

\begin{figure}[ht]
%\begin{minipage}[t]{1.4in}
% \vspace{-0.00cm}
\centering 
  \includegraphics[width=1.00\columnwidth]{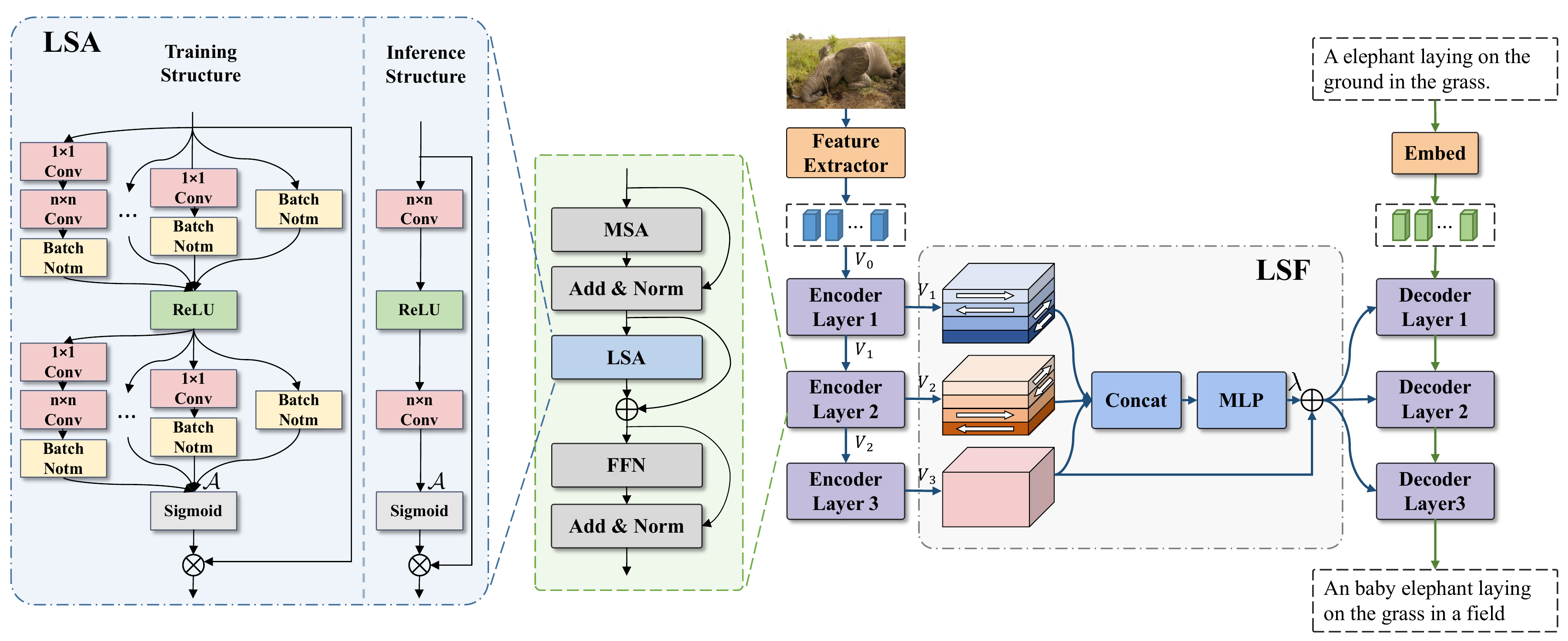}
  \vspace{-0.00cm}

{\color{black}{
  \caption{Overview of our proposed LSTNet. Grid features are used as visual representations and fed to the visual encoder.  Locality-Sensitive Attention (LSA) is inserted after self-attention to capture local dependencies. We also apply reparameterization {\color{black}{technology}} to reduce its overhead during inference.   The {\color{black}{output}} of encoding layers {\color{black}{is}} further processed by Locality-Sensitive Fusion (LSF) to achieve cross-layer aggregation, which can also provide the semantics of different layers for local visual modeling. MSA and FFN represent Multi-head Self-Attention and Feed-Forward Networks, respectively.
  }
  \label{fig:fig2}
}}

  \vspace{-0.00cm}
\end{figure}

\subsection{Overview}
As shown in Fig.~\ref{fig:fig2}, our proposed \emph{Locality-Sensitive Transformer Network (LSTNet)}  follows the encoder-decoder paradigm. Concretely, the encoder takes the visual features as input and then models their relationships by the encoder layers, where \emph{Locality-Sensitive Attention (LSA)} module is adopted to enhance local visual modeling. Then, \emph{Locality-Sensitive Fusion (LSF)} aggregates the visual features from different encoding layers, based on which the decoder predicts caption words to describe the given visual content. 

The visual features before the $l$-th encoder layer are denoted as $V_{l-1} \in \mathbb{R}^{N_v \times c}$ ($N_v = h \times w$), where $h$, $w$, $c$ represent the height, width and channel dimension of  visual features, respectively.

Each encoder layer of LSTNet consists of three components: (1) a Multi-head Self-Attention (MSA) module; (2) a Locality-Sensitive Attention (LSA) module; (3) a Feed-Forward Network (FFN). The visual features $V_{l-1}$ from the last encoder layer is first processed by the MSA as follows (LayerNorm operation is omitted for conciseness):
\begin{equation}
    V^{\prime}_{l-1} = V_{l-1} + MSA(V_{l-1}, V_{l-1}, V_{l-1}),
\label{eq:SA1}
\end{equation}
where $MSA(\cdot)$ is the standard Multi-head Self-Attention in Transformer \cite{vaswani2017attention}. {\color{black} Because MSA can model the relationship and context of any two positions in the input sequence, MSA is conducive to capturing long-range dependencies and modeling global information among grids.}

Self-Attention is often inefficient in capturing the local details, which, however, is critical for grid visual features as explained in Sec.\ref{sec:intro}. Thus, based on $V^{\prime}_{l-1}$, LSA is adopted to capture  the  dependencies of neighboring grids to further refine visual features:
\begin{equation}
    V^{\prime \prime}_{l-1} = V^{\prime}_{l-1} + LSA(V^{\prime}_{l-1}),
    \label{eq:LSA}
\end{equation}
where the detail of LSA is described in the next subsection. {\color{black}{Because LSA is composed of cascaded convolution layers, which can model the relationship between adjacent grids, LSA is conducive to modeling local relationships among grids.}} 

Then the output of the LSA module is fed to FFN for the interaction in the channel domain:
\begin{equation}
    V_l = V^{\prime \prime}_{l-1} + FFN(V^{\prime \prime}_{l-1}),
\label{eq:FFN1}
\end{equation}
\begin{equation}
    FFN(x) = \max ( 0 , x W_1 + b_1) W_2 + b_2.
\label{eq:FFN2}
\end{equation}

Different from previous Transformer-based models, which only feed the output of the top encoder layer to the decoder, our proposed LSF module aggregates  visual features from all encoder layers to obtain richer features in semantic by Locality-Sensitive Fusion (LSF):
\begin{equation}
    V^* = LSF(V_1, V_2, \cdots , V_n),
    \label{eq:FFN}
\end{equation}
where $n$ is the number of encoder layers. 

Finally, $V^*$ is fed into the decoder to generate the captions, which is the same as that of the vanilla Transformer \cite{vaswani2017attention}.

\subsection{Locality-Sensitive Attention (LSA)}
As visualized in Fig.~\ref{fig:fig1}(b), an object in the image may be divided into several fragments and distributed in various grids, which destroys the spatial and semantic information of visual objects. A reasonable approach is to strengthen the interaction of local information, which is also in line with the assumption that features close to each other in vision are more likely to be correlated.  

Thus, to capture the local details and model the interaction between adjacent grids, we propose a multi-scale locality-sensitive module, namely Locality-Sensitive Attention (LSA). Specifically, {\color{black}{the output feature $V^{\prime} \in \mathbb{R}^{N \times C}$ of the MSA module is a grid sequence, where $N$ is the number of grids, $C$ is the size of channel dimension. We first reshape $V^{\prime} \in \mathbb{R}^{ N\times C}$ to $V^{\prime} \in \mathbb{R}^{ H \times W \times C}$, where $H, W$ are the height and width of the grid feature.}} {\color{black}{Then, we  use two multi-scale 2D CNNs in series with an activation function (\emph{i.e.,} ReLU) in between to obtain the visual features $\mathcal{A}$ after multi-scale local perception, which could be formulated as follows:}}
\begin{equation}
    \mathcal{A} = MSC_2\big(\sigma(MSC_1(V^{\prime})\big),
    \label{eq:att}
\end{equation}
where $\sigma(\cdot)$ is the activation function and $MSC_i(\cdot)$  represents a multi-scale CNN implemented by multi-branch CNNs: 
\begin{equation}
    MSC_i(x) =  BN_1^i\left(F_1^i(x)\right)  + \cdots +BN_N^i\left(F_N^i(x)\right),
    \label{eq:cnn}
\end{equation}
where $i \in \{1, 2\}$, $N$ is the number of branches, $BN_j(\cdot)$ is Batch Normalization \cite{ioffe2015batch}, $F_j(\cdot)$ represents identity mapping, one convolution module, or several  convolution modules in series, and $j \in \{1, \cdots, N\}$. In our LSTNet, the number of branches $N$ is 3. {\color{black}{As shown in the blue area in Fig.~\ref{fig:fig2}, three branches are (1) the identity mapping, (2) the $1 \times 1$ Conv, and (3) the sequential combination of $1 \times 1$ Conv and $3 \times 3$ Conv, respectively. }}

During inference, the  multi-branch structure $MSC_i(\cdot)$ can be simplified into a single-branch structure to save the number of parameters and computational cost by using some structural reparameterization techniques \cite{ding2021repvgg,ding2021diverse} without any performance loss:
\begin{equation}
    MSC_i(x) \rightarrow F^i(x),
    \label{eq:cnn2}
\end{equation}
where $F^i(x)$ is a $3 \times 3$ convolution, and $i \in \{1, 2\}$. 

{\color{black}{To get the attention weight for each grid, we also apply the Sigmoid function to $\mathcal{A}$.}} Finally, we reweight the output feature of Self-Attention layer $V^{\prime}$ according to locality-sensitive attention map as follows:
{\color{black}{
\begin{equation}
    V^{\prime \prime}=V^{\prime} \otimes Sigmoid(\mathcal{A}),
    \label{eq:sig}
\end{equation}
}}
where  {\color{black}{$\otimes$}} represents element-wise multiplication.

\subsection{Locality-Sensitive Fusion (LSF)}
 Features from different layers tend to contain semantic information of various levels \cite{cornia2020meshed}. However, most existing image captioning methods only feed the feature of the top encoder layer to the decoder, leading to {\color{black}{low-level information loss}}. To avoid such information loss, we fuse the features of all layers in the encoder and then feed the fused feature into the decoder. 
 
 Technically, we introduce a simple \emph{Spatial  Shift} operation to enable each grid to align with its neighbor grids, and then the Multi-Layer Perceptron (MLP) is used to interact not only in the channel domain but also in the spatial domain.
 
\begin{figure}[ht]

%\begin{minipage}[t]{1.4in}
\centering 
  \includegraphics[width=0.7\columnwidth]{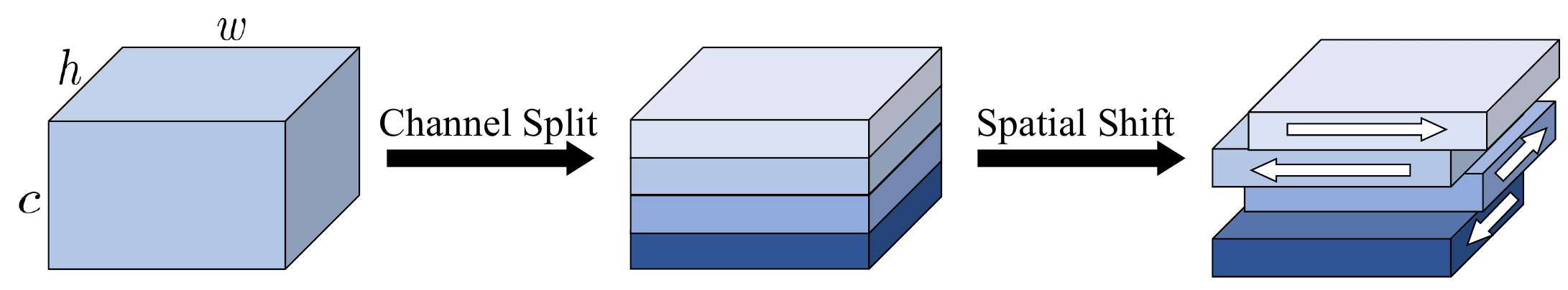}
  \vspace{-0.00cm}
  \caption{ Illustration of the \emph{Spatial Shift} operation. $c,h,w$ are the size of {\color{black}{the channel}}, height, and width dimension, respectively.
  }
  \label{fig:spatialshift}
\end{figure}

 Particularly, denoting features from the $l$-th encoder layer as $V_l \in \mathbb{R}^{h \times w \times c}$ (the reshape operation is omitted here),  $V_1$ and $V_2$ are shifted by different Spatial  Shift operations (\emph{i.e.,} $SS_1(\cdot)$ and $SS_2(\cdot)$), {\color{black}{which can be represented as Eq. \ref{eq:SS1} and Eq. \ref{eq:SS2}:

% \begin{small}
\begin{equation}
    \begin{array}{l}
    V_1 [d_s : h, :, 0 : c/4]  = V_1 [0 : h - d_s, :, 0 : c/4],\\
    V_1 [0 : h - d_s, :, c/4  : c/2] = V_1 [d_s : h, :, c/4  : c/2],\\
    V_1 [:, d_s : w, c/2 : 3c/4]  =  V_1 [:, 0 : w  - d_s, c/2 : 3c/4],\\
    V_1 [:, 0 : w - d_s, 3c/4 : c]  =  V_1 [:, d_s : w, 3c/4 : c],
    \end{array}
\label{eq:SS1}
\end{equation}
\begin{equation}
    \begin{array}{l}
    V_2 [:, d_s : w, 0  : c/4]  = V_2 [:, 0 : w - d_s, 0  : c/4],\\
    V_2 [:, 0 : w - d_s, c/4 : c/2]  =  V_2 [:, d_s : w, c/4 : c/2], \\
    V_2 [d_s : h, :, c/2 : 3c/4 ] = V_2 [ 0 : h - d_s, :, c/2 : 3c/4],\\
    V_2 [0 : h - d_s, :, 3c/4 : c] = V_2 [d_s : h, :, 3c/4 : c],\\
    \end{array} 
\label{eq:SS2}
\end{equation}
% \end{small}
where $V_i$ is the output feature of the $i$-th encoder layer, $d_s$ is the shift distance of Spatial Shift, which determines the scope of local interaction. The output of the top encoder layer $V_3$ is not processed by any shift operations. The illustration of Spatial Shift can be observed in Fig.~\ref{fig:fig2} and Fig.~\ref{fig:spatialshift}.
 
 }}

\begin{table}[]
    \small
	\centering
	{\color{black}{
	\caption{Leaderboard of the published state-of-the-art image captioning models on the COCO online testing server.}
	\label{tab:online_sota}
	}}

% 	\vspace{-0.00cm}
	\resizebox{1.00\columnwidth}{!}{
	
	{\color{black}{
	\begin{tabular}{l|cccccccccccccc}
		\hline
		\multirow{2}{*}{Model}    & \multicolumn{2}{c}{BLEU-1}   & \multicolumn{2}{c}{BLEU-2}   & \multicolumn{2}{c}{BLEU-3}   & \multicolumn{2}{c}{BLEU-4}   & \multicolumn{2}{c}{METEOR}   & \multicolumn{2}{c}{ROUGE-L}  & \multicolumn{2}{c}{CIDEr-D}    \\ 
		    & c5            & c40           & c5            & c40           & c5            & c40           & c5            & c40           & c5            & c40           & c5            & c40           & c5             & c40            \\ \hline
		
		SCST      \cite{rennie2017self} \tiny{cvpr'17}     & 78.1          & 93.7          & 61.9          & 86.0          & 47.0          & 75.9          & 35.2          & 64.5          & 27.0          & 35.5          & 56.3          & 70.7          & 114.7          & 116.0          \\ 
		LSTM-A    \cite{yao2017boosting} \tiny{iccv'17}    & 78.7          & 93.7          & 62.7          & 86.7          & 47.6          & 76.5          & 35.6          & 65.2          & 27.0          & 35.4          & 56.4          & 70.5          & 116.0          & 118.0          \\ 
		Up-Down    \cite{anderson2018bottom} \tiny{cvpr'18}  & 80.2          & 95.2          & 64.1          & 88.8          & 49.1          & 79.4          & 36.9          & 68.5          & 27.6          & 36.7          & 57.1          & 72.4          & 117.9          & 120.5          \\ 
		RFNet   \cite{jiang2018recurrent} \tiny{eccv'18}   & 80.4          & 95.0          & 64.9          & 89.3          & 50.1          & 80.1          & 38.0          & 69.2          & 28.2          & 37.2          & 58.2          & 73.1          & 122.9          & 125.1          \\ 
		GCN-LSTM    \cite{yao2018exploring}  \tiny{eccv'18} & 80.8             & 95.2             & 65.5          & 89.3          & 50.8          & 80.3          & 38.7          & 69.7          & 28.5          & 37.6          & 58.5          & 73.4          & 125.3          & 126.5          \\ 
		SGAE   \cite{yang2019auto} \tiny{cvpr'19}     & 81.0          & 95.3          & 65.6          & 89.5          & 50.7          & 80.4          & 38.5          & 69.7          & 28.2          & 37.2          & 58.6          & 73.6          & 123.8          & 126.5          \\ 
		AoANet   \cite{huang2019attention} \tiny{cvpr'19}  & 81.0          & 95.0          & 65.8          & 89.6          & 51.4          & 81.3          & 39.4          & 71.2          & 29.1          & 38.5          & 58.9          & 74.5 & 126.9          & 129.6          \\ 
  
		CAVPN \cite{zha8684270} \tiny{TPAMI'19} & 80.1  &94.9  &64.7 & 88.8 & 50.0 & 79.7  &37.9  &69.0 & 28.1  &37.0  &58.2 & 73.1 & 121.6 & 123.8  \\
		%GCN-HIP  & \textbf{81.6} & \textbf{95.9} & \textbf{66.2} & 90.4          & \textbf{51.5} & 81.6          & \textbf{39.3} & 71.0          & 28.8          & 38.1          & \textbf{59.0} & 74.1          & 127.9          & 130.2          \\ \hline
		%USUA     & 79.9          & 94.7          & 64.3          & 88.6          & 49.5          & 79.3          & 37.4          & 68.3          & 28.2          & 37.1          & 57.9          & 72.8          & 123.1          & 125.5          \\ \hline
		ETA   \cite{li2019entangled}  \tiny{iccv'19}     & 81.2          & 95.0          & 65.5          & 89.0          & 50.9          & 80.4          & 38.9          & 70.2          & 28.6          & 38.0          & 58.6          & 73.9          & 122.1          & 124.4          \\
		$M^2$Transformer    \cite{cornia2020meshed}  \tiny{cvpr'20}  & 81.6         & 96.0          & 66.4          & 90.8          & 51.8          & 82.7          & 39.7          & 72.8          & 29.4          & 39.0          & 59.2         & 74.8          & 129.3          & 132.1          \\
		XTransformer  (ResNet-101)   \cite{pan2020x} \tiny{cvpr'20}   & 81.3          & 95.4          & 66.3          & 90.0          & 51.9          & 81.7          & 39.9          & 71.8          & 29.5          & 39.0          & 59.3          & 74.9          & 129.3          & 131.4          \\ 
		XTransformer  (SENet-154)   \cite{pan2020x} \tiny{cvpr'20}  &81.9  &95.7  &66.9  &90.5  &52.4  &82.5  &40.3  &72.4  &29.6  &39.2  &59.5  &75.0  &131.1  &133.5\\
		
        DLCT (ResNeXt101) \cite{luo2021dual} \tiny{aaai'21}  &82.0 & 96.2 & 66.9 & 91.0 & 52.3 & 83.0 & 40.2 & 73.2 & 29.5 & 39.1 & 59.4 & 74.8 & 131.0 & 133.4\\
        DLCT (ResNeXt152) \cite{luo2021dual}  \tiny{aaai'21} & 82.4 & 96.6 & 67.4 & 91.7 & 52.8 & 83.8 & 40.6 & 74.0 & 29.8 & 39.6 & 59.8 & 75.3 & 133.3 & 135.4\\
		
		RSTNet(ResNext101)  \cite{zhang2021rstnet}  \tiny{cvpr'21} &81.7  &96.2  &66.5  &90.9 & 51.8 & 82.7 & 39.7 & 72.5 & 29.3 & 38.7 & 59.2 & 74.2 & 130.1 & 132.4 \\

		RSTNet(ResNext152) \cite{zhang2021rstnet} \tiny{cvpr'21} & 82.1 & 96.4 & 67.0 & 91.3 & 52.2 & 83.0 & 40.0 & 73.1 & 29.6 & 39.1 & 59.5 & 74.6 & 131.9 & 134.0\\

		{\color{black}{GAT \cite{WANG2022117174} \tiny{expert syst. appl.'22}}} & 81.1& 95.1 & 66.1 & 89.7 & 51.8 & 81.5 & 39.9 &71.4 &29.1 & 38.4 & 59.1 & 74.4 &127.8 &129.8 \\
		{\color{black}{VPNet \cite{WANG202248} \tiny{neurocomputing'22}}} & 81.4& 	95.5	& 66.4	& 90.3	& 52.0	& 82.1	& 40.0	& 72.1& 	29.3& 	38.9	& 59.3	& 74.9	& 128.2	& 130.6\\

        \color{black}{CtxAdpAtt~\cite{zhang2021exploring} \tiny{tmm'22}} & 81.0 & 95.2 & 65.5 &91.0 & 51.5 &81.7 & 39.3 & 70.9 & 29.4 & 39.0 & 59.6 & 75.1 & 128.5 & 131.0 \\
      
      \color{black}{ReFormer~\cite{yang2022reformer} \tiny{mm'22}} &  82.0& 96.7&-&-&-&-& 40.1& 73.2& 29.8& 39.5& 59.9& 75.2& 129.9& 132.8\\

		 \hline
		LSTNet  (ResNeXt-101)  & 82.2&96.2&67.2&91.2&52.7&83.5&40.6&73.8&29.6&39.3&59.6&75.0&132.0&134.5 \\
		
		LSTNet  (ResNeXt-152)     & \textbf{82.6}      & \textbf{96.7} & \textbf{67.8}  & \textbf{92.0} & \textbf{53.3 } & \textbf{84.3} & \textbf{41.1} & \textbf{74.7} & \textbf{29.9} & \textbf{39.6} & \textbf{60.0} & \textbf{75.4}          & \textbf{134.0} & \textbf{136.3} \\
		\hline

	\end{tabular}

	}}

}
    \vspace{-0.00cm}

	\vspace{-0.00cm}
\end{table}

\begin{table}[t]
	{\color{black}{
	\caption{Comparisons with SOTAs on the Karpathy test split. All values are reported as {\color{black}{percentages}}  (\%),  where B-N,  M,  R, and  C   are short for BLEU-N,  METEOR,  ROUGE-L, and  CIDEr scores.}
	\label{tab:offline_sota}
	}}
\centering

\begin{small}

% \resizebox{1.00\columnwidth}{!}{

{\color{black}{

	\begin{tabular}{l|c c c c c c}
	\hline
	    Model        &B-1  & B-4& M & R & C & S \\ 
	\hline
    SCST  \cite{rennie2017self}   \tiny{cvpr'17}     &   -  & 34.2 & 26.7 & 55.7 & 114.0&  -   \\ %\cline{2-8}
     Up-Down  \cite{anderson2018bottom}  \tiny{cvpr'18}     & 79.8 & 36.3 & 27.7 & 56.9 & 120.1& 21.4 \\ %\cline{2-8}
     RFNet  \cite{jiang2018recurrent}   \tiny{eccv'18}      & 79.1 & 36.5 & 27.7 & 57.3 & 121.9& 21.2 \\ %\cline{2-8}
     GCN-LSTM  \cite{yao2018exploring}   \tiny{eccv'18}   & 80.5 & 38.2 & 28.5 & 58.3 & 127.6& 22.0 \\ %\cline{2-8}
     SGAE  \cite{yang2019auto}   \tiny{cvpr'19}      & 80.8 & 38.4 & 28.4 & 58.6 & 127.8& 22.1 \\ %\cline{2-8}
     
      CAVPN \cite{zha8684270} \tiny{tpami'19} &- & 38.6 & 28.3 & 58.5 & 126.3 & 21.6 \\
      % ETA  \cite{li2019entangled}         & \textbf{81.5} & \textbf{39.3} & 28.8 & \textbf{58.9} & 126.6& 22.7 \\ %\cline{2-8}
    % LBPF         & 80.5 & 38.3 & 28.5 & 58.4 & 127.6& 22.0 \\ \hline
     AoANet  \cite{huang2019attention}  \tiny{cvpr'19}    & 80.2 & 38.9 & 29.2 & 58.8 & 129.8& 22.4 \\ %\cline{2-8}
     ORT  \cite{herdade2019image}   \tiny{neurips'19}       & 80.5 & 38.6 & 28.7 & 58.4 & 128.3& 22.6 \\ %\cline{2-8}
     Transformer  \cite{vaswani2017attention} \tiny{neurips'17}  & 80.7 & 38.6 &29.1  & 58.5 & 130.1& 22.7 \\ %\cline{2-8}
     $M^2$Transformer  \cite{cornia2020meshed}  \tiny{cvpr'20} & 80.8 & 39.1 & 29.2 & 58.6 & 131.2& 22.6 \\ %\cline{2-8} 
     XTransformer  \cite{pan2020x} \tiny{cvpr'20}  & 80.9 & 39.7 & 29.5 & 59.1 & 132.8& 23.4 \\ 

    DLCT \cite{luo2021dual} \tiny{aaai'21}  & 81.4& 39.8 &29.5&59.1& 133.8 & 23.0 \\

      RSTNet  \cite{zhang2021rstnet}  \tiny{cvpr'21} & 81.1 & 39.3& 29.4& 58.8& 133.3 &23.0 \\

      \color{black}{GAT \cite{WANG2022117174} \tiny{expert syst. appl.'22}} & 80.8 & 39.7 & 29.1 & 59.0 & 130.5 &22.9 \\ 
      
      \color{black}{VPNet \cite{WANG202248} \tiny{neurocomputing'22}} & 80.9	& 39.7	& 29.3	& 59.2	& 130.4	& 23.2 \\
      
      \color{black}{CtxAdpAtt~\cite{zhang2021exploring} \tiny{tmm'22}} & 80.5 & 39.1 & 29.3 & 59.3 & 130.1 & \textbf{23.6} \\
      
      \color{black}{ReFormer~\cite{yang2022reformer} \tiny{mm'22}} & - & 39.8 & 29.7 & 59.8 & 131.2 & 23.0 \\

      \hline %\cline{2-8} 

        %   \hline %\cline{2-8} 
          
         LSTNet &  \textbf{81.5}          & \textbf{40.3} & \textbf{29.6} & \textbf{59.4} & \textbf{134.8} & 23.1  \\
        \hline
        \end{tabular}

}}

% }
    \end{small}
    \vspace{-0.00cm}

    % \vspace{-0.00cm}
\end{table}

Then, shifted features from different layers are concatenated together:
\begin{equation}
    V_c=Concat(V_1,V_2,V_3).
\label{eq:concat}
\end{equation}

{\color{black}{
Theoretically,  MLP can't model the relationships among adjacent grids. However, after being shifted by the Spatial Shift operation, which aligns each grid with its neighbors, MLP can communicate in both channel and spatial domains:
}}
\begin{equation}
    \tilde{V}=\sigma(V_cW_1)W_2,
\label{eq:mlp}
\end{equation}
where $\sigma(\cdot)$ is the ReLU activation function, $W_1 \in \mathbb{R}^{3c \times 3c}$ and $W_2 \in \mathbb{R}^{3c \times c}$ are the learnable projection matrices.

To further enhance the descriptive power of visual features, we combine the outputs of the top encoder layer with the fused feature $\tilde{V}$ via a residual connection:
\begin{equation}
    V^*= \lambda \tilde{V} +V_{top},
\label{eq:add}
\end{equation}
where $V_{top}$ is the feature of the top encoder layer,  $V_{top} = V_3$ in our LSTNet, and $\lambda$ serves as a weighting factor.

{\color{black}{
Specifically, the motivation {\color{black}{for}} LSF comes from two aspects:
1) The output feature map of different encoder layers has different semantics (\emph{i.e.,} the high-level feature map has high-level semantic information, and the low-level feature map has low-level semantic information). The traditional Transformer only feeds the feature map of the last layer in the encoder into the decoder, ignoring the low-level semantic information. LSF solves this problem by fusing the outputs of all encoder layers, considering both high-level and low-level semantic information.
2) The LSF module can interact with local grids through spatial shift operation, which is conducive to modeling object-level information.
}}

\subsection{Objectives}
Given the ground-truth caption $y^*_{1:T}$ and the captioning model with parameters $\theta$, where $T$ is the length of the caption, we pre-train the model using Cross-Entropy (CE) loss as follows:

\begin{equation}
    L_{CE}=-\sum_{t=1}^T \log\left(p_\theta(y^*_t|y^*_{1:t-1})\right).
\label{eq:ce}
\end{equation}

Then we further optimize the model using CIDEr and BLEU-4 scores by Self-Critical Sequence Training (SCST) \cite{rennie2017self} as follows:
\begin{equation}
    \nabla_\theta L_{RL}(\theta) = - \frac{1}{k} \sum^k_{i=1}\left(r(y^i_{1:T}) - b\right)\nabla_\theta \log p_\theta(y^i_{1:T}),
\label{eq:grid}
\end{equation}
where $k$ is the beam size, $r(\cdot)$ is the sum of CIDEr and BLEU-4, and $b = \left(\sum_i r(y^i_{1:T})\right)/k$ is the reward baseline.

\section{Experiment}
\subsection{Datasets}
We conduct our experiments on the popular MS-COCO \cite{lin2014microsoft} image captioning dataset. It contains 123,287 images, including 82,783 training images, 40,504 validation images, and 40,775 testing images, each of which is annotated with 5 captions. We adopt the split provided by \cite{karpathy2015deep} for the offline test, where 5,000 images are used for validation, 5,000 images for testing, and the rest images for training. Besides, we also upload generated captions of the official testing set for online evaluation \footnote{\url{https://competitions.codalab.org/competitions/3221\#results}}.

\subsection{Implementation Details}
{\color{black}{The grid features are extracted from a pre-trained Faster-RCNN \cite{ren2016faster} provided by \cite{jiang2020defense}, where a stride-1 $C_5$ backbone and 1 × 1 RoIPool with
two FC layers are used as the detection head {\color{black}{for training}} Faster R-CNN on the VG dataset. Specifically, we adopt the $C_5$ feature maps and average-pool them as $7 \times 7$ spatial size.}} Note that we do not use any extra data preprocessing, except simple augmentations (\emph{e.g.,} RandomCrop, RandomRotation). The $d_{model}$ in the LSTNet is 512, the expansion rate in the FFN is 4, the number of heads is 8, and the size of the beam search is 5. 

We use Adam optimizer to train our model in both stages and adopt the relative position encoding following \cite{luo2021dual}. In the CE training stage, the batch size is 50, and the learning rate is linearly increased to $1 \times 10^{\text{-}4}$ during the first 4 epochs. Afterwards, we set it to $2 \times 10^{\text{-}5}$, $4 \times 10^{\text{-}6}$ at 10-th and 12-th epoch. After 18 epochs of CE pre-training, we optimize the model by SCST with the batch size of 100 and learning rate of $5 \times 10^{\text{-}6}$. The learning rate will be set to $2.5 \times 10^{\text{-}6}$, $5 \times 10^{\text{-}7}$, $2.5 \times 10^{\text{-}7}$, $5 \times 10^{\text{-}8}$ at the 35-th, 40-th, 45-th, 50-th epoch, and the SCST training will last 42 epochs.

\subsection{Performance Comparison}

In this section, we compare our LSTNet with SOTAs on both offline and online {\color{black}{evaluations}}. The compared models include: SCST  \cite{rennie2017self}, Up-Down  \cite{anderson2018bottom}, RFNet  \cite{jiang2018recurrent}, GCN-LSTM  \cite{yao2018exploring}, SGAE  \cite{yang2019auto},  AoANet  \cite{huang2019attention}, ETA \cite{li2019entangled}, ORT  \cite{herdade2019image}, Transformer  \cite{vaswani2017attention}, $M^2$Transformer  \cite{cornia2020meshed}, XTransformer  \cite{pan2020x}, RSTNet  \cite{zhang2021rstnet} and DLCT \cite{luo2021dual}. Following the standard evaluation criterion, we adopt BLEU-N \cite{papineni2002bleu}, METEOR \cite{banerjee2005meteor}, ROUGE-L \cite{lin2004rouge}, CIDEr \cite{vedantam2015cider}, SPICE \cite{anderson2016spice} to evaluate the performance.
% Note that the reason why the pre-training models (\emph{e.g.,} Oscar \cite{li2020oscar}, M6 \cite{lin2021m6}) are not included is that too much additional data is exposed to them (\emph{e.g.,} 6.5 million text-image pairs for  Oscar, 1.9 TB images and 292 GB texts for M6), leading to a unfair comparison.

\subsubsection{Online Evaluation}
% \noindent\textbf{Online Evaluation.}
Tab.~\ref{tab:online_sota} shows the performance comparisons of LSTNet and other SOTA methods on the online COCO test server with 5 reference captions (c5) and 40 reference captions (c40). For fair comparisons, we also use the ensemble of four models following \cite{cornia2020meshed} and adopt two common backbones (\emph{i.e.,} ResNeXt-101, ResNeXt-152 \cite{xie2017aggregated}). Notably, our LSTNet outperforms other SOTA methods in all metrics by significant margins. Surprisingly, we observe that \emph{LSTNet with \textbf{ResNeXt-101}} performs better than \emph{RSTNet with \textbf{ResNeXt-152}} and \emph{X-Transformer with \textbf{SENet-154}} on most metrics.

\subsubsection{Offline Evaluation}
% \noindent\textbf{Offline Evaluation.}
Tab.~\ref{tab:offline_sota} summarizes the performance of the state-of-the-art models and our approach {\color{black}{to}} the offline COCO Karpathy test split. Note that for fair comparisons, we report the results of single models without using any ensemble technologies. We can observe that our proposed LSTNet outperforms all the other SOTA models in terms of most metrics.  Notably, the CIDEr score of our LSTNet achieves \textbf{\emph{134.8\%}}, outperforming the strongest competitor DLCT by \textbf{\emph{1.0\%}}, which adopts both region and grid features. 
{\color{black}{

We obverse that the LSTNet with the grid feature performs better than some models (\emph{e.g.,} $M^2$Transformer \cite{cornia2020meshed}, XTransformer \cite{pan2020x}) with region features. We think the reasons why the grid-level scheme in our paper performs better than the {\color{black}{object-level}} scheme are as follows:
1) The region feature's background information is missing, and the grid feature extracts all of the information in the image. Specifically, visual region features are collected from the image's salient parts, typically omitting contextual information. Because of the lack of background information, the model performs poorly in capturing relationships between objects. The grid feature, on the other hand, collects all spatial information from the image.
2) The pre-trained object detector often involves noisy, overlapped, or erroneous detections, which ultimately limits the performance upper bound of image captioning models. On the other hand, {\color{black}{grid features}} do not provide detection information, so the impact of error detection is avoided.
3) In grid features, an object is divided into different grids, and this is the motivation of this paper. Our approach enables the model to capture local information and solves this problem.

}}

\subsubsection{Fair Comparisons with SOTA Methods}

\begin{table}[t]
	{\color{black}{
	\caption{ Comparisons with SOTA methods on the Karpathy test split using the same ResNeXt101 \cite{xie2017aggregated} grid feature. }
	\label{tab:grid}
	}}

\centering
	\small
% 	\resizebox{1.00\columnwidth}{!}{
{\color{black}{
	\begin{tabular}{l|c c c c c c}
		%\toprule
		\hline
	    Model        &B-1  & B-4& M & R & C & S \\ 
	    \hline
	    
		Up-Down  \cite{anderson2018bottom} \tiny{cvpr'18}    & 75.0 & 37.3 &28.1  & 57.9 & 123.8& 21.6 \\ %\cline{2-8} 
         AoANet  \cite{huang2019attention}  \tiny{cvpr'19}      & 80.8 & 39.1 &29.1  & 59.1 & 130.3& 22.7 \\ %\cline{2-8}
 		 Transformer  \cite{vaswani2017attention} \tiny{neurips'17}  & 81.0 & 38.9 &29.0  & 58.4 & 131.3& 22.6 \\ %\cline{2-8} 
 		 $M^2$Transformer \tiny{cvpr'20}  \cite{cornia2020meshed} & 80.8 & 38.9 &29.1  & 58.5 & 131.8& 22.7 \\ %\cline{2-8}
        XTransformer  \cite{pan2020x} \tiny{cvpr'20}  & 81.0 & \textbf{39.7} &29.4  & 58.9 & 132.5& \textbf{23.1} \\
        DLCT \cite{luo2021dual}  \tiny{aaai'21}  & 81.4 & 39.8 & 29.5 & 59.1 & 133.8 & 23.0 \\   %\cline{2-8} 
        RSTNet  \cite{zhang2021rstnet}  \tiny{cvpr'21}  & 81.1 & 39.3& 29.4& 58.8& 133.3 &23.0 \\  \hline %\cline{2-8} 

         LSTNet &  \textbf{81.5}          & \textbf{40.3} & \textbf{29.6} & \textbf{59.4} & \textbf{134.8} & \textbf{23.1}         \\
		
		\hline
	\end{tabular}
}}
    % 	}
% 	\vspace{-0.3cm}

% 	\vspace{-0.3cm}
\end{table}

To eliminate the interference of different visual features, we conduct experiments on the same grid features to compare the LSTNet and other SOTA methods. As reported in Tab.~\ref{tab:grid},  compared with other methods on the same visual features, our proposed LSTNet still achieves superior performance on all metrics.

\subsection{Ablation Study}

\subsubsection{Effect of Different Branches of LSA}
% \noindent\textbf{Effect of Different Branches of LSA.}
To validate the impact of each branch, we conduct a series of experiments by leveraging different branches of LSA. The performance of LSA with different branches is illustrated in Tab.~\ref{tab:branch}. {\color{black}{
By analyzing this table, we gain the following observations:
\begin{itemize}
    \item Compared with the model without LSA (line 1), adopting LSA (line 2-7) with either one or more branches is helpful to generate better captions. Moreover, the more branches are adopted, the better performance tends to be achieved. This may be because the proposed LSA module improves the local perception ability of the model, so it is beneficial to the perception of object information.

    \item As shown in Tab. \ref{tab:branch}, equipping one branch (lines 2,3,4) outperforms the model with 0 {\color{black}{branches}} (line 1) on most evaluation metrics. By comparing the model with one branch (lines 2,3,4) to the model with two branches (lines 5,6,7), we can see that the model with two branches outperforms the model with one branch on most metrics, particularly CIDEr. Furthermore, we can see that the completed LSTNet (line 8) with all three branches performs the best. This may be attributed to that objects in the image vary in size, so more branches are conducive to {\color{black}{strengthening}} the multi-scale modeling ability for {\color{black}{objects of}} different sizes. Importantly, by using reparameterization techniques, more branches of LSA will not lead to higher overhead during inference.
\end{itemize}

}}

\subsubsection{Effect of Different Arrangements of LSA and SA}
% \noindent\textbf{Effect of Different Arrangements of LSA and SA}
To explore the impact of various arrangements of Locality-Sensitive Attention (LSA) and Self-Attention (SA), we compare three methods to combine LSA and SA: (1) sequential LSA-SA, (2) sequential SA-LSA, (3) parallel usage of SA and LSA. As shown in Tab.~\ref{tab:arrangement}, we can observe that the performance of the sequential SA-LSA is better than the others. {\color{black}{The main reason may be that the features processed by SA are coarse-grained, and our proposed LSA, which aims to model local relationships, is helpful to further refine the visual features.}}

\begin{figure}[ht]
% \vspace{-0.4cm}
\centering 
  \includegraphics[width=0.75\columnwidth]{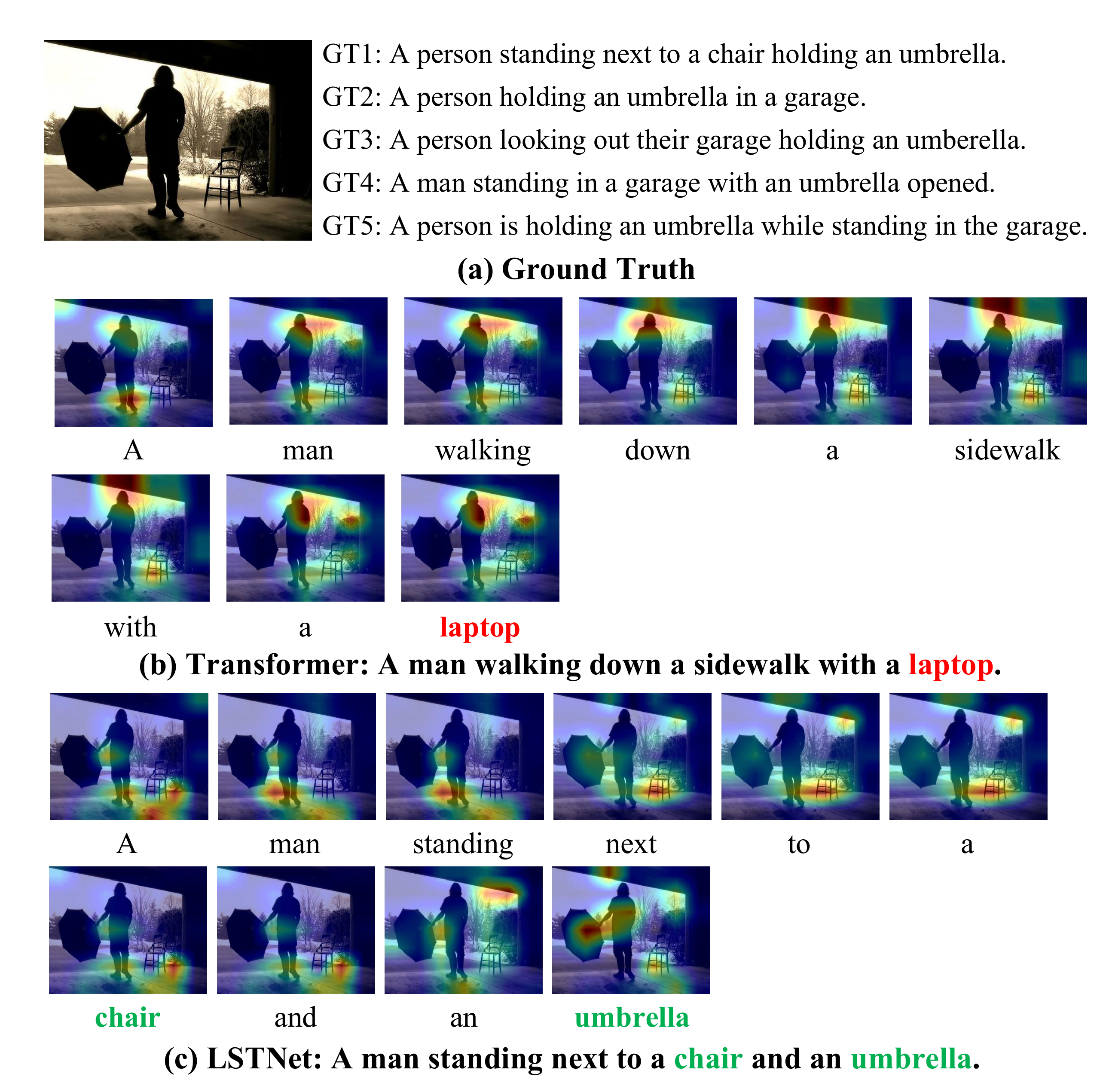}
%   \vspace{-0.3cm}
{\color{black}{
  \caption{ The ground truth (a) and the visualization of attended images along with the caption generation of standard Transformer (b) and  LSTNet (c). The most significant difference between the two generated captions is highlighted in bold.}
  \label{fig:fig5}
  }}
  
%   \vspace{-0.3cm}
\end{figure}

\begin{figure}[ht]
%\begin{minipage}[t]{1.4in}
\centering 
  \includegraphics[width=0.55\columnwidth]{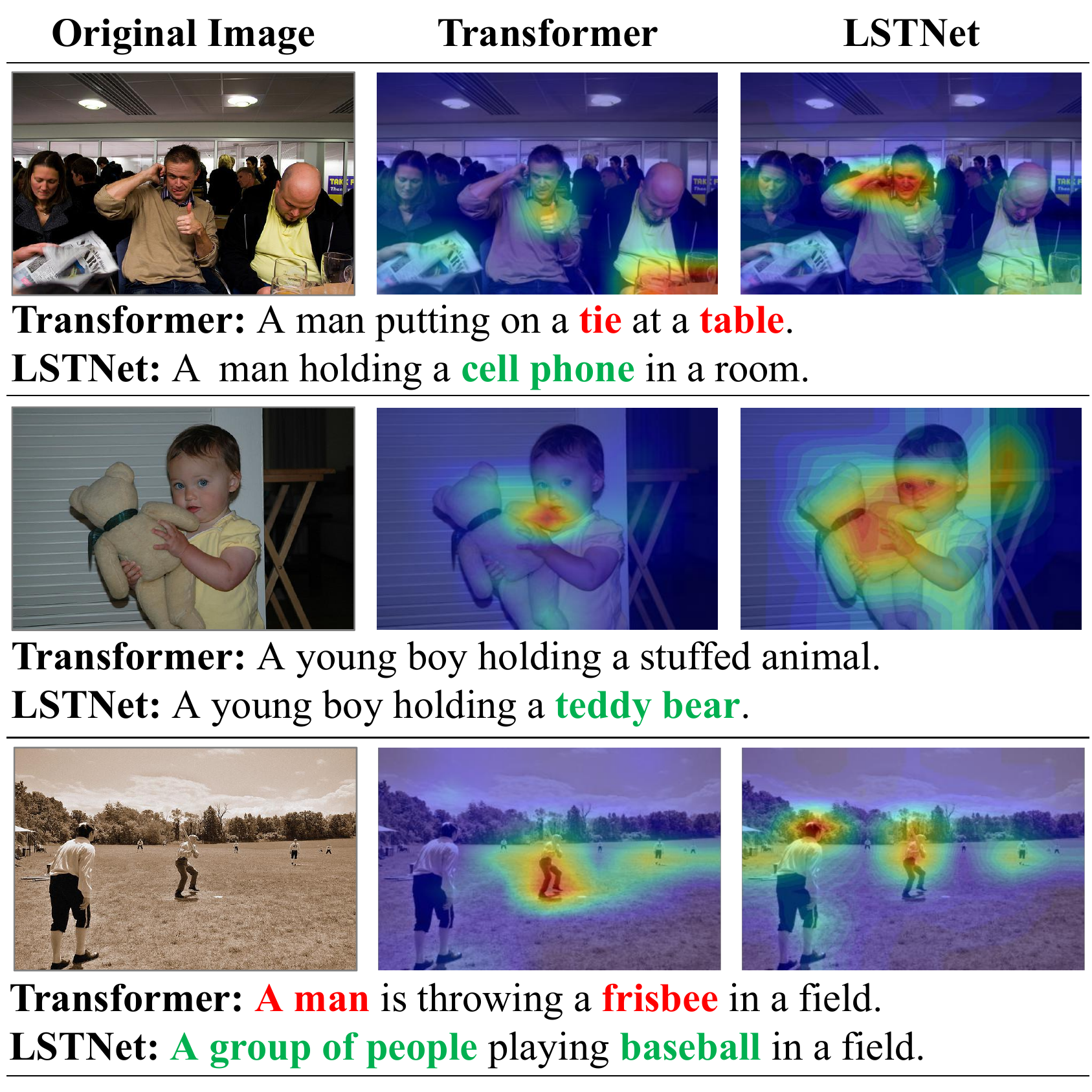}
  \vspace{-0.00cm}
  \caption{ Some examples of captions generated by Transformer and LSTNet on the same grid features and comparisons of  corresponding attention map of the top encoder layer of Transformer and LSTNet.
  }
  \label{fig:fig3}
\end{figure}

\begin{table}[t]
	{\color{black}{
\caption{Ablation studies on various branches. All branches contain BatchNorm, and all models are not equipped with LSF. All values are reported as {\color{black}{percentages}}  (\%),  B-N,  M,  R,  C, and S are short for BLEU-N,  METEOR,  ROUGE-L,  CIDEr, and SPICE scores.}
\label{tab:branch}
	}}

\centering
% \small
% \resizebox{1.00\columnwidth}{!}{
\begin{tabular}{lll|cccccc}

\hline
Identity      & 1$\times$1      & {\color{black}{1$\times$1+3$\times$3}}       & B-1           & B-4           & M             & R             & C              & S             \\
\hline
\textbf{$\times$} & \textbf{$\times$} & \textbf{$\times$} & 81.0          & 38.9          & 29.0          & 58.4          & 131.3          & 22.6          \\
\textbf{$\surd$}  & \textbf{$\times$} & \textbf{$\times$} & 81.1          & 38.9          & 29.1          & 58.4          & 131.6          & 22.6          \\
\textbf{$\times$} & \textbf{$\surd$}  & \textbf{$\times$} & 81.1          & 39.2          & 29.2          & 58.9          & 132.2          & 22.7          \\
\textbf{$\times$} & \textbf{$\times$} & \textbf{$\surd$}  & \textbf{81.2} & 39.4          & 29.2          & 58.9          & 132.3          & 22.7          \\
\textbf{$\surd$}  & \textbf{$\surd$}  & \textbf{$\times$} & \textbf{81.2} & 39.6          & 29.1          & 59.0          & 133.5          & 22.7          \\
\textbf{$\surd$}  & \textbf{$\times$} & \textbf{$\surd$}  & \textbf{81.2} & 39.6          & 29.2          & 58.9          & 133.4          & 22.6          \\
\textbf{$\times$} & \textbf{$\surd$}  & \textbf{$\surd$}  & \textbf{81.2} & 39.6          & \textbf{29.3} & \textbf{59.1} & 133.5          & \textbf{22.8} \\
\hline
\textbf{$\surd$}  & \textbf{$\surd$}  & \textbf{$\surd$}  & \textbf{81.2} & \textbf{39.7} & \textbf{29.3} & \textbf{59.1} & \textbf{133.6} & \textbf{22.8}\\
\hline
\end{tabular}
% }
\vspace{-0.00cm}

\vspace{-0.00cm}
\end{table}

\begin{table}[t]
	{\color{black}{
\caption{Ablation studies on various arrangements of SA and LSA.  + is sequential connection, \& represents parallel connection.}
\label{tab:arrangement}
	}}

\centering
% \resizebox{1.00\columnwidth}{!}{
\begin{tabular}{l|cccccc}
\hline
      Arrangement           & B-1           & B-4           & M             & R             & C              & S             \\
\hline
LSA + SA  & 81.1          & 39.6          & 29.1          & 59.0          & 133.4          & \textbf{22.8} \\
SA + LSA  & \textbf{81.2} & \textbf{39.7} & \textbf{29.3} & \textbf{59.1} & \textbf{133.6} & \textbf{22.8} \\
SA \& LSA & 81.0          & 39.5          & 28.9          & 59.0          & 132.7          & 22.7        \\
\hline
\end{tabular}
% }
% \vspace{-0.3cm}

% \vspace{-0.3cm}
\end{table}

\begin{table}[t]
	{\color{black}{
\caption{ Ablation studies on various shift distances of LSF. All models are equipped with both LSA and LSF modules.}
\label{tab:dis}
	}}

\centering
\small
% \resizebox{1.00\columnwidth}{!}{
\begin{tabular}{l|cccccc}
\hline
Shift Distance & B-1           & B-4           & M             & R             & C              & S             \\
\hline
$d_s$ = 0            & 81.2          & 39.9          & 29.3          & 59.1          & 133.9          & 22.8          \\
$d_s$ = 1      &       \textbf{81.5}          & \textbf{40.3} & \textbf{29.6} & \textbf{59.4} & \textbf{134.8} & \textbf{23.1}  \\
$d_s$ = 2            & 81.4          & 40.1          & 29.5 & 59.2          & 134.4          & 22.9          \\
$d_s$ = 3            & 81.3          & 40.1          & 29.5 & 59.2          & 134.2          & 22.8          \\
$d_s$ = 4            & 81.3          & 40.0          & 29.4          & 59.1          & 134.0          & 22.8       \\
\hline
\end{tabular}
% }
\vspace{-0.00cm}

\vspace{-0.00cm}
\end{table}

\begin{table}[t]
	{\color{black}{
\caption{ Performance comparisons with different weight factors $\lambda$. All models are equipped with both LSA and LSF modules.}
\label{tab:lambda}
	}}

% \resizebox{1.00\columnwidth}{!}{
\centering
\begin{tabular}{l|cccccc}
 \hline
$\lambda$ & B-1           & B-4           & M             & R             & C              & S             \\ \hline
$\lambda$ = 0.1       & \textbf{81.7} & \textbf{40.3} & 29.4          & 59.3          & 134.3          & 22.9          \\
$\lambda$ = 0.2       & 81.5          & \textbf{40.3} & \textbf{29.6} & \textbf{59.4} & \textbf{134.8} & \textbf{23.1} \\
$\lambda$ = 0.3       & 81.6          & 40.2          & 29.5          & 59.3          & 134.5          & 23.0          \\
$\lambda$ = 0.5       & 81.6          & \textbf{40.3} & 29.5          & 59.3          & 134.5          & 30.0          \\
$\lambda$ = 0.7       & 81.4          & 40.0          & 29.4          & 59.0          & 134.0          & 22.9       \\   \hline  
\end{tabular}
% }
\vspace{-0.00cm}

% \vspace{-0.00cm}
\end{table}

\begin{table}[t]
	{\color{black}{
\caption{Ablation studies on LSA and LSF modules.}
\label{tab:abla}
	}}

\centering
\small
% \resizebox{1.00\columnwidth}{!}{
\begin{tabular}{l|cccccc}
\hline
Module    & B-1           & B-4           & M             & R             & C              & S             \\
\hline
w/o LSA+LSF          & 81.0          & 38.9          & 29.0          & 58.4          & 131.3          & 22.6 \\
only LSA  & 81.2          &39.7         & 29.3    & 59.1        & 133.6       & 22.8 \\
only LSF  & 81.3 & 39.8        &   29.3 & 59.0        & 133.7        & 22.8 \\ \hline
LSA + LSF &  \textbf{81.5}          & \textbf{40.3} & \textbf{29.6} & \textbf{59.4} & \textbf{134.8} & \textbf{23.1} \\
\hline
\end{tabular}
% }
\vspace{-0.00cm}

\vspace{-0.00cm}
\end{table}

\begin{table}[t]
	{\color{black}{
\caption{Ablation studies on various fusion methods. All models are not equipped with LSA.  FPN is the  feature pyramid network similar to \cite{liu2018path}. All values are reported as {\color{black}{percentages}}  (\%),  where B-N,  M,  R,  C, and S are short for BLEU-N,  METEOR,  ROUGE-L,  CIDEr, and SPICE scores.}
\label{tab:fuse}
	}}

\centering
\small
% \resizebox{1.00\columnwidth}{!}{
\begin{tabular}{l|cccccc}
\hline
Fusion methods    & B-1           & B-4           & M             & R             & C              & S             \\
\hline
w/o Fuse          & 81.0          & 38.9          & 29.0          & 58.4          & 131.3          & 22.6          \\
MLP               & 81.0          & 39.3          & 29.2          & \textbf{59.0}          & 132.5          & 22.7          \\
SumPool           & 81.1          & 39.4          & 29.1          & \textbf{59.0}          & 132.5          & 22.7          \\
3 $\times$ 3 Conv & \textbf{81.3} & 39.7         & 29.0          & 58.9          & 132.6          & 22.7          \\
FPN               & 81.1          & 39.1          & 29.1          & 58.7          & 132.6          & 22.6          \\ \hline
LSF (ours)        & \textbf{81.3} & \textbf{39.8} & \textbf{29.3} & \textbf{59.0} & \textbf{133.7} & \textbf{22.8}\\
\hline
\end{tabular}
% }

\end{table}

\begin{table}[]
	{\color{black}{
\caption{Ablation studies on different grid sizes of the visual feature. All values are reported as percentages  (\%),  where B-N,  M,  R,  C and S are short for BLEU-N,  METEOR,  ROUGE-L,  CIDEr, and SPICE scores.}
\label{tab:gridsize}
	}}
\centering

{\color{black}{
\begin{tabular}{l|cccccc}
\hline
Grid Size                 & B-1  & B-4  & M    & R    & C     & S    \\ \hline
$1 \times 1$ & 78.5 & 36.2 & 27.4 & 56.5 & 120.3 & 20.6 \\
$2 \times 2$ & 79.9 & 37.7 & 28.3 & 57.7 & 124.9 & 21.7 \\
$3 \times 3$ & 80.4 & 38.9 & 28.8 & 58.4 & 129.6 & 22.3 \\
$4 \times 4$ & 80.6 & 38.9 & 28.9 & 58.5 & 130.3 & 22.4 \\
$5 \times 5$ & 81.0 & 39.6 & 29.1 & 58.8 & 131.7 & 22.7 \\
$6 \times 6$ & 81.3 & 39.9 & 29.3 & 58.9 & 132.1 & 22.9 \\
$7 \times 7$ & 81.5 & 40.3 & 29.6 & 59.4 & 134.8 & 23.1 \\ \hline
\end{tabular}
}}
\end{table}

\subsubsection{Effect of Different Shift Distances of LSF}
% \noindent\textbf{Effect of Different Shift Distances of LSF.}
To explore the impact of shift distance $d_s$ of LSF, we conduct experiments by increasing $d_s$ from 0 to 4 ($d_s = 0$ means that all features are not shifted). From Tab.~\ref{tab:dis}, we could observe that shifted LSF performs better than the unshifted one. {\color{black}{This can be attributed to that shifted LSF makes each grid can interact with neighboring grids during fusing, thus enhancing the local modeling.}} However, when the shift distance is greater than 1, the performance begins to drop and $d_s = 1$ performs best. {\color{black}{The reason may be that the large shift promotes the long-distance interaction, but ignores local modeling.}}

\subsubsection{Effect of $\lambda$ in LSA}
% \noindent\textbf{Effect of $\lambda$ in LSA.}
To choose the best weighting factor $\lambda$  in Eq.~\ref{eq:add}, we also conduct a group of experiments. From Tab.~\ref{tab:lambda}, we could find that too large $\lambda$ will lead to performance degradation, and $\lambda = 0.2$ performs well on most metrics. Thus, we use $\lambda = 0.2$ for our experiments if not specified.

\subsubsection{Effect of Decoupling LSA and LSF}
% \noindent\textbf{Effect of Decoupling LSA and LSF.}
To gain insights into the proposed LSA and LSF modules, we decouple these two modules in the experiment. As reported in Tab.~\ref{tab:abla}, compared with the complete model, {\color{black}{the}} performance of the model without LSA+LSF degrades dramatically. Particularly, it drops absolutely by \textbf{\emph{1.4\%}} and \textbf{\emph{3.5\%}} on BLEU-4 and CIDEr respectively, which demonstrates the vital importance of LSA and LSF. Particularly, our LSA and LSF achieve performance gains of 2.3\% and 2.4\% on the CIDEr score, respectively. {\color{black}{This shows that LSA and LSF modules can promote each other to achieve better performance.}}

\begin{table}
	{\color{black}{
\caption{Performance comparisons of different captioning metrics for the Standard Transformer and our LSTNet.  P-values come from two-tailed t-tests using paired samples. P-values in bold are significant at  0.05  significance level.}
\label{tab:p_metric}
	}}

\centering
%\vspace{-0.5em}
\begin{center}
\resizebox{1.0\columnwidth}{!}{
\begin{tabular}{l|cccccc}
\toprule
Model       & BLEU-1               & BLEU-4            & METEOR        & ROUGE & CIDEr & SPICE \\
\midrule
Transformer & 81.0            & 38.9          & 29.0            & 58.4  & 131.3 & 22.6  \\
LSTNet  & 81.5            & 40.3            & 29.6            & 59.4            & 134.8           & 23.1            \\ \hline
p-value & {$\mathbf{6.7 \times 10^{-3}}$}  & {$\mathbf{3.3 \times 10^{-7}}$} & {$\mathbf{1.2 \times 10^{-7}}$} & {$\mathbf{8.0 \times 10^{-7}}$} & {$\mathbf{3.9 \times 10^{-9}}$} & {$\mathbf{1.2 \times 10^{-4}}$}\\
\bottomrule
\end{tabular}
}

\end{center}
%\vspace{-0.5em}
\end{table}

\begin{table}
	{\color{black}{
\caption{ Subcategories of SPICE metrics for the Standard Transformer and our proposed LSTNet. P-values are calculated by two-tailed t-tests using paired samples. Note that p-values in bold are significant at  0.05  significance level. }
\label{tab:p_spice}
	}}

\centering
\small

%\vspace{-0.5em}
% \resizebox{1.00\columnwidth}{!}{
    \begin{tabular}{l|cccccc}
    \toprule
    \multirow{2}{*}{Model} & \multicolumn{6}{c}{SPICE}                                  \\ \cline{2-7}
       & Relation & Cardinality & Attribute & Size & Color & Object \\ 
       \hline
      Transformer  &6.91           & 20.58          & 11.80          & 4.71  & 12.93          & 40.35             \\
LSTNet  & 7.06  & 22.68 & 12.38          & 4.84           & 14.60          & 40.76          \\ \hline
p-value & 0.298 & 0.059 & \textbf{0.002} & \textbf{0.048} & \textbf{0.001} & \textbf{0.009} \\

    \bottomrule
    \end{tabular}
% }

% \vspace{-1.5em}
\end{table}

\subsubsection{Effect of Different Approaches to Fuse Features}
To justify the effectiveness of LSF, we design a group of experiments by replacing LSF with various modules to fuse features.  As shown in Tab. \ref{tab:fuse}, we can find that fusing features from different layers improves the performance when compared with the results in the first row  (\emph{i.e,} without fusing features).{\color{black}{ The main reason may be that features from different layers are complementary in semantic information and fusing features will enrich visual features.}} Moreover, our proposed LFS module outperforms other modules by a large margin, which strongly illustrates the validity of LFS.

\subsubsection{\textcolor{black}{Effect of Different Grid Sizes}}
%越大越好
%因为越大，包含的视觉细节越多，越有助于生成更加细节的caption

To explore the impact of different grid {\color{black}{sizes}}, we conduct a series of experiments by setting  the visual features to different sizes via average pooling. As shown in Tab. \ref{tab:gridsize}, the performance of image captioning gradually improves as the grid size is increased. {\color{black}{This is explained by the fact that a larger image feature offers more fine-grained and richer semantic information, and the image captioning model will produce more accurate descriptions as a result of these details.}}

\subsection{Quantitative Analysis}

By comparing conventional metrics (\emph{e.g.,} BLEU-N, CIDEr, SPICE), it is difficult to determine whether our method significantly improves the performance of image captioning. Aiming to demonstrate the efficacy and superiority of our proposed LSTNet in {\color{black}{an intuitive way}}, we conduct a two-tailed t-test with paired samples to compare LSTNet with {\color{black}{a}} standard Transformer. To be specific, we first perform the two-tailed t-test for each conventional metric to explore whether the quality of caption generated by LSTNet is significantly improved compared with the standard Transformer. Besides, we also report the semantic subcategories of SPICE scores (\emph{i.e.,} Relation, Cardinality, Attribute, Size, Color, and Object), which can be used to measure the semantic relevance between generated sentences and ground truth. Furthermore, for each comprehensive SPICE score, we do a detailed two-tailed t-test with matched data to see if these semantic indicators have also been significantly improved.

The conventional metrics and corresponding p-values for {\color{black}{the}} t-test over the test set are displayed in Tab. \ref{tab:p_metric}. We observe that the improvement of all metrics is statistically significant under the significant level $\alpha = 0.05$, which {\color{black}{demonstrates}} that our proposed LSTNet is conducive to the quality of the generated caption. Tab. \ref{tab:p_spice} details the semantic subcategories of SPICE scores and p-values for t-test over the test set. We can observe that all semantic subcategories of SPICE are improved, which reveals the effectiveness and superiority of local visual modeling in LSTNet. Furthermore, we can also observe that under the significant level $\alpha = 0.05$, some semantic subcategories of SPICE (\emph{i.e.,} Attribute, Size, Color, and Object) obtain significant improvements. Notably, all these four metrics describe the attributes of the object, so it proves that the local visual modeling in LSTNet is helpful to capture object-level information. Compared with other semantic metrics, the improvement of \emph{Relation} metric is relatively insignificant. This may be because, to capture the relation between objects, it is not enough to only improve the local modeling ability of the model and global modeling ability is also very important. For other semantic metrics of a single object (\emph{i.e.,} Attribute, Size, Color, and Object), local visual modeling has been able to achieve significant performance improvement.

\subsection{Qualitative Analysis}

To qualitatively validate the effectiveness of LSTNet, we display several typical examples of captions generated by Transformer and LSTNet on the same grid features in Fig.~\ref{fig:fig3}. We can observe that the captions generated by Transformer are {\color{black}{uninformative}} even erroneous, while the captions generated by LSTNet are more accurate and distinguishable, which demonstrates that our proposed LSA and LSF are helpful to recognize the visual object by local modeling.

To gain deep insights into the reason why LSTNet can generate accurate captions, we further illustrate the attention map of the top encoder layer in Transformer and LSTNet in Fig.~\ref{fig:fig3}. By analyzing the results, we gain the following observations: 1) The attention map produced by Transformer fails to attend to the important visual objects in the image, while LSTNet is able to focus on the important ones. For instance, for the image in the first row in Fig.~\ref{fig:fig3}, Transformer is focusing on the table and LSTNet is focusing on the man and phone. Thus, Transformer generates the erroneous caption (\emph{i.e.,} \emph{``tie''}, \emph{``table''}), while LSTNet recognizes \emph{``A man holding a cell phone''} correctly. 2) Transformer can only attend to one object or small area in the image, while LSTNet will focus on more primary objects, thus generating accurate and detailed descriptions. For example, for the images in the second row in Fig.~\ref{fig:fig3}, Transformer is only focusing on the mouth of the boy, and LSTNet is focusing on both the boy and the teddy bear. Thus, Transformer fails to recognize the \emph{ ``teddy bear''} but only produces a general phrase (\emph{i.e.,} \emph{``a stuffed animal''}). Thanks to the precise attention in the encoder, LSTNet recognizes \emph{``a young boy holding a teddy bear ''}  successfully. These observations reveal that our proposed LSA forces the model to focus on not only important but also comprehensive information in the image.

\begin{table}[]
	{\color{black}{
\caption{Comparison with {\color{black}{state of the art}} on the Flickr8k dataset. All values are reported as {\color{black}{percentages}}  (\%),  where B-N,  M,  R, and  C   are short for BLEU-N,  METEOR,  ROUGE-L, and  CIDEr scores. † indicates an ensemble model results.}
\label{tab:flickr8k}
	}}

\centering
{\color{black}

% \resizebox{1.00\columnwidth}{!}{
\begin{tabular}{l|ccccc}
\toprule
Methods        & B-1            & B-4            & M             & R             & C             \\ \midrule
Deep VS \cite{karpathy2015deep}       & 57.9          & 16.0          & -             & -             & -             \\
Google NIC \cite{vinyals2015show}†   & 63.0          & -             & -             & -             & -             \\
Soft-Attention \cite{xu2015show} & 67.0          & 19.5          & 18.9          & -             & -             \\
Hard-Attention \cite{xu2015show} & 67.0          & 21.3          & 20.3          & -             & -             \\
emb-gLSTM  \cite{jia2015guiding}    & 64.7          & 21.2          & 20.6          & -             & -             \\
Log Bilinear  \cite{donahue2015long}  & 65.6          & 17.7          & 17.3          & -             & -             \\ \hline
LSTNet          & \textbf{67.4} & \textbf{24.3} & \textbf{21.5} & \textbf{44.8} & \textbf{63.6} \\ \bottomrule
\end{tabular}
% }
}

\end{table}

\subsection{ Attention Visualization}

To better qualitatively evaluate the generated results with LSTNet, we visualize the contribution of each grid of the visual features during caption generation in Fig.~\ref{fig:fig5}. Technically, we average attention weights of 8 heads in the last decoder layer. We can observe that Transformer will attend to  irrelevant regions, thus generating erroneous descriptions (\emph{e.g.,} \emph{``laptop''}).  Instead, our proposed LSTNet can focus on correct grids when generating informative words like \emph{``chair''} and \emph{``umbrella''}. These observations demonstrate that our proposed LSA and LSF modules help the model consistently focus on the correct regions for image captioning by providing richer and finer-grained visual features for the decoder through local interaction and fusion.

\subsection{ Generalization on the Flickr Datasets}

\begin{table}[]
	{\color{black}{
\caption{Comparison with {\color{black}{state of the art}} on the Flickr30k dataset. All values are reported as {\color{black}{percentages}}  (\%),  where B-N,  M,  R, and  C  are short for BLEU-N,  METEOR,  ROUGE-L and  CIDEr scores. † indicates an ensemble model results.}
\label{tab:flickr30k}
	}}

\centering
{\color{black}

% \resizebox{1.00\columnwidth}{!}{
\begin{tabular}{l|ccccc}
\toprule
Methods        & B1            & B4            & M             & R             & C             \\ \midrule
Deep VS   \cite{karpathy2015deep}      & 57.3          & 15.7          & -             & -             & -             \\
Google NIC \cite{vinyals2015show}†    & 66.3          & 18.3          & -             & -             & -             \\
m-RNN  \cite{mao2014deep}        & 60.0          & 19.0          & -             & -             & -             \\
Soft-Attention \cite{xu2015show} & 66.7          & 19.1          & 18.5          & -             & -             \\
Hard-Attention \cite{xu2015show} & 66.9          & 19.9          & 18.5          & -             & -             \\
emb-gLSTM   \cite{jia2015guiding}    & 64.6          & 20.6          & 17.9          & -             & -             \\
ATT \cite{you2016image}†           & 64.7          & 23.0            & 18.9          & -             & -             \\
Log Bilinear  \cite{donahue2015long}  & 60.0          & 17.1          & 16.9          & -             & -             \\ \hline
LSTNet          & \textbf{67.1} & \textbf{23.3} & \textbf{20.4} & \textbf{44.3} & \textbf{64.5} \\ \bottomrule
\end{tabular}
% }
}

\end{table}

To verify the generalization of our proposed LSNet, we also conduct extensive experiments on the Flickr8k and Flickr30k {\color{black}{datasets}}.
%\subsubsection{\textbf{Performance Comparison on Flickr8k}}
%We first conduct experiments on the classical Flickr8k dataset.  
The performance comparisons between our proposed LSTNet and previous SOTAs on Flickr8k \cite{hodosh2013framing} and Flickr30k \cite{young2014image} are shown in Tab \ref{tab:flickr8k} and Tab~\ref{tab:flickr30k}, respectively. As can be observed, our proposed LSTNet outperforms the previous SOTAs with a significant margin on both Flickr8k and Flickr30K. This verifies that our proposed LSTNet has strong generalization on other datasets.

\section{Conclusion}

In this paper, we {\color{black}{propose}} LSTNet, a novel Locality-Sensitive Transformer Network for image captioning, which exploits the local interaction and fusion for better object recognition {\color{black}{with grid features}}. {\color{black}{Specifically,}} we design the LSA module to model the {\color{black}{relationship}} within {\color{black}{an encoder layer}}, which is helpful to attain the complete semantic information of each object and refine the details of visual features. We then introduce the LSF module to fuse visual features from different layers by modeling the relationships between adjacent grids, which leads to visual features with richer semantic information. Experimental results on the MS-COCO dataset demonstrate the significant performance advantages of  LSTNet over previous SOTA models. Extensive ablation studies and visualization comparisons further reveal the effectiveness and insights of the individual components of LSTNet.  Additional experiments on the Flickr8k and Flickr30k datasets also {\color{black}{verify}} the generalization of LSTNet on other datasets. {\color{black} Although the proposed LSTNet significantly improves the object location capability and the captioning performance, it also introduces additional computation and parameters. However, in comparison to the self-attention of the vanilla Transformer, the parameters and computation introduced by LSA and LSF modules are negligible. }



 \bibliographystyle{elsarticle-num} 
 \bibliography{pr22}

\end{document}